# MSNGO：multi-species protein function annotation based on 3D protein structure and network propagation

Beibei Wang[1], Boyue Cui[1], Shiqu Chen[1], Xuan Wang[1, 4], Yadong Wang[2, 3] and Junyi Li[1, 3, 4] *

[1] School of Computer Science and Technology, Harbin Institute of Technology (Shenzhen), Shenzhen, Guang Dong 518055, China.

[2] Center for Bioinformatics, faculty of computing, Harbin Institute of Technology, Harbin, Heilongjiang 150001, China

[3] Key Laboratory of Biological Bigdata, Ministry of Education, Harbin Institute of Technology, Harbin, Heilongjiang 150001, China

[4] Guangdong Provincial Key Laboratory of Novel Security Intelligence Technologies, Harbin Institute of Technology (Shenzhen), Shenzhen, Guangdong 518055, China

*To whom correspondence should be addressed.

**Abstract**
**Motivation:** In recent years, protein function prediction has broken through the bottleneck of sequence features, significantly improving prediction accuracy using high-precision protein structures predicted by AlphaFold2. While single-species protein function prediction methods have achieved remarkable success, multi-species protein function prediction methods are still in the stage of using PPI networks and sequence features. Providing effective cross-species label propagation for species with sparse protein annotations remains a challenging issue. To address this problem, we propose the MSNGO model, which integrates structural features and network propagation methods. Our validation shows that using structural features can significantly improve the accuracy of multi-species protein function prediction.
**Results:** We employ graph representation learning techniques to extract amino acid representations from protein structure contact maps and train a structural model using a graph convolution pooling module to derive protein-level structural features. After incorporating the sequence features from ESM-2, we apply a network propagation algorithm to aggregate information and update node representations within a heterogeneous network. The results demonstrate that MSNGO outperforms previous multi-species protein function prediction methods that rely on sequence features and PPI networks.
**Availability:** https://github.com/blingbell/MSNGO.
**Contact:** lijunyi@hit.edu.cn
**Supplementary information:** Supplementary data are available at *Bioinformatics* online.

## 1 Introduction

Protein function prediction is essential for understanding cellular processes and mechanisms. Accurate function annotation not only clarifies protein roles in pathways but also provides insights into uncharacterized proteins. With the rapid growth of genomics and proteomics data, efficient computational methods are urgently needed for large-scale protein function prediction. Experimental verification is inefficient, resource-intensive, and cannot meet current demands. In contrast, computational methods greatly reduce prediction time and costs. Therefore, it is urgent to efficiently and accurately predict protein functions through bioinformatics.

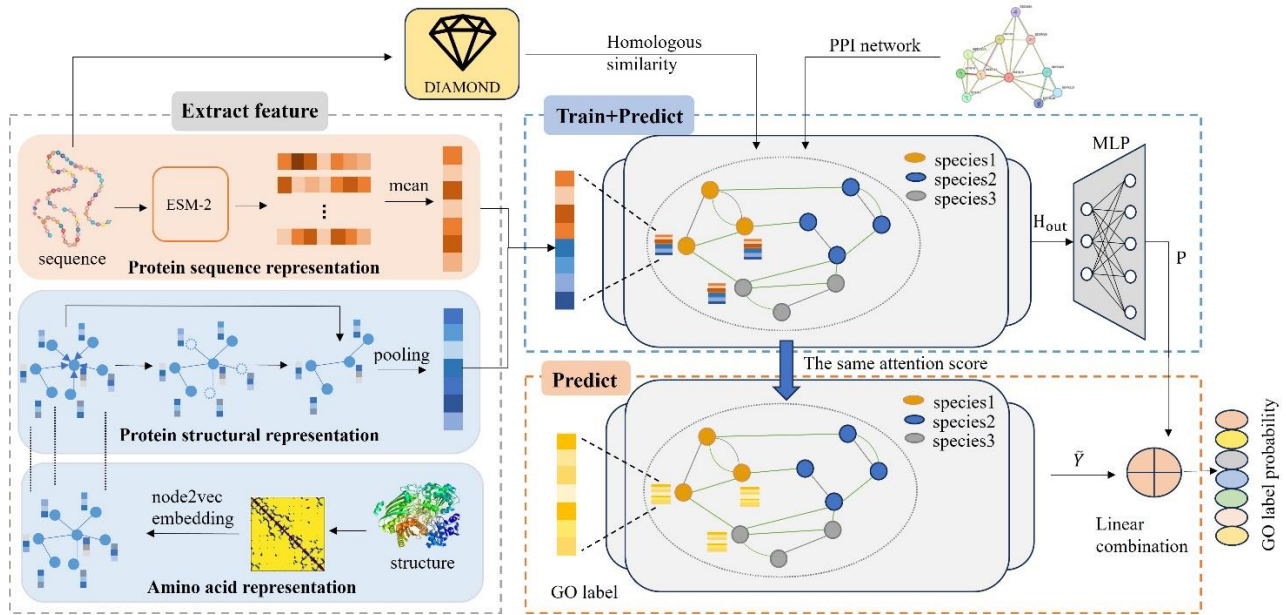

**Fig. 1. An overview of MSNGO model.** The model inputs protein sequences, structures, and PPI networks. During feature extraction, a contact map is built from the 3D structure using Cα atom distances, and node2vec provides amino acid-level representations. The structural model based on graph convolution pooling modules extracts hidden vectors as protein structural features, while ESM-2 generates sequence features by averaging amino acid representations. The structural representation and sequence representation are spliced into the final feature representation of a protein. In training, the concatenated sequence and structural features are propagated through a heterogeneous network of PPI and homology similarity networks, with an MLP mapping the final representations to GO terms. During prediction, GO label vectors and concatenated protein features are propagated, with label propagation sharing attention scores from feature propagation. The final prediction is a linear combination of feature propagation and label propagation.

To facilitate the description of protein functions, Gene Ontology (Ashburner, et al., 2000) standardizes the functions of genes and their products into three domains: Biological Process (BP), Molecular Function (MF), and Cellular Component (CC). Consequently, the problem of protein function prediction is abstracted into a multi-label classification problem. In recent years, bioinformatics has made significant progress in the field of protein function prediction, particularly with methods that have excelled in the CAFA (Dessimoz, et al., 2013) competition. These methods often rely on multiple data sources to capture more comprehensive information, resulting in better classification performance.

This success is largely attributed to the precise structural predictions made possible by models like AlphaFold2 (Jumper, et al., 2021), which have resolved the issue of scarce structural data. Therefore, AlphaFold2's open structural database (Varadi, et al., 2022) undoubtedly offers a breakthrough in overcoming the reliance on sequence features for protein function prediction. Currently, many methods utilize the protein structures predicted by AlphaFold2 as datasets. PredGO (Zheng, et al., 2023) embeds three-dimensional structural features using the GVP-GNN (Jing, et al., 2020) model and integrates sequence features and PPI networks to achieve high-performance protein function prediction. DeepFRI (Gligorijevic, et al., 2021) leverages a graph convolution-based model to train on protein structure contact maps, going beyond homology transfer by capturing both local sequence features and global structural characteristics. Struct2GO (Jiao, et al., 2023) employs hierarchical graph pooling with an attention mechanism to pool amino acid-level representations, resulting in protein-level representations. Not limited to structure feature, multimodal biological data continues to grow. DPFunc (Wang, et al., 2025) uses domain information for GO prediction. MIF2GO (Ma, et al., 2024) employs fusion technology to integrate six biological data modalities, achieving a robust protein representation.

Most of the mentioned prediction techniques are based on single-species studies. Therefore, they require training different models for multi-species data, leading to increased time and hardware resource consumption. Additionally, there is an issue of data imbalance between species. For some species with sparse protein function labels, transferring annotations is a significant challenge. Limited by this issue, multi-species protein function prediction techniques have only recently begun to emerge. These techniques aim to predict the functions of proteins across different species simultaneously, significantly reducing the time required for multi-species function prediction and addressing the issue of sparse labels in target species.

Among existing methods, DeepGraphGO (You, et al., 2021) was the first to attempt expanding from single-species prediction models to the multi-species domain. It uses GCN (Kipf and Welling, 2016) to simultaneously convolve the PPI networks of multiple species, achieving the goal of predicting protein functions across multiple species simultaneously. However, the prediction results rely solely on the propagation outcomes within the PPI networks of each species. For species with sparse labels, the final classification results do not effectively guide the identification of potential protein functions. On the same dataset, the NetQuilt (Barot, et al., 2021) model was proposed, which integrates sequence similarity and PPI networks by calculating IsoRank (Singh, et al., 2008) similarity scores, resulting in a multi-species protein isomorphic network. Although this method significantly improves the performance, it fundamentally relies on network alignment to find cross-species connections, and the number of protein nodes input is limited by the size of the GPU. To improve the efficiency of network propagation, S2F (Torres, et al., 2021) introduced a new label propagation algorithm. By linearly combining different transition networks, it can effectively learn information from various levels within isomorphic networks. SPROF-GO (Yuan, et al., 2023) made improvements on the label propagation algorithm proposed by S2F. This model constructs multi-species networks using homology information, which allows for the rapid prediction of GO labels for newly discovered proteins. Unlike the propagation method in S2F, which is based on isomorphic networks, PSPGO (Wu, et al., 2022) proposes a heterogeneous network propagation method. It constructs multi-species heterogeneous networks using sequence homology networks and PPI networks.

Our proposed model, MSNGO, leverages homology similarity and PPI networks to achieve cross-species functional propagation and incorporates structural data to make protein-level representations more complete. MSNGO has three notable features: First, it achieves cross-species label propagation through a multi-species heterogeneous network. We construct a heterogeneous network based on sequence homology similarity and PPI networks, using an attention mechanism to update node vectors and propagate labels. Second, it introduces structural data into multi-species network propagation for the first time. The structural data comes from the public database of AlphaFold2, which achieves atomic-level accuracy. Using the predicted three-dimensional structures from AlphaFold2 significantly improves the accuracy of function prediction. Third, it employs a model based on graph convolution and hierarchical graph pooling to learn structural features suitable for network propagation. The trained model extracts embedded representations of the structural data, allowing structural features to propagate through large, complex networks.

## 2 Methods and materials

### 2.1 Overview

Inspired by PSPGO, our model uses homology similarity and PPI networks to form a heterogeneous network for propagation. As shown in Figure 1, the process is divided into three main steps:

(1) **Feature Extraction Stage:** We learn effective feature vectors for protein sequences and structures separately. Protein sequences are processed using the ESM-2 (Rives, et al., 2021) model, while a structural model is trained for protein structures. The hidden vectors extracted from the structural model serve as the protein structural features.

(2) **Training Stage:** Protein structural and sequence features are concatenated to form node feature vectors in the heterogeneous network. Each network learns an attention score matrix to measure node closeness during propagation. Node vector updates depend on attention scores, layer weights, and the vectors themselves. Through multi-layer propagation, the model learns effective protein features from both networks, and an MLP performs multi-label classification.

(3) **Prediction Stage:** In the heterogeneous network, both protein features and label vectors are propagated to proteins with unknown labels. Label vectors from the training set propagate, allowing test set proteins to aggregate effective labels. The results of label and feature propagation are then linearly combined and fed into the MLP classifier.

### 2.2 Datasets

We collected datasets from 13 species, including mammals, ectotherms, plants, bacteria, and fungi. We present the statistics of the dataset in Table S1 in the Supplementary Data. Data includes PPI networks, GO annotations, Gene Ontology files, protein sequences, and structures. PPI networks are from STRING (v11.0b) (Szklarczyk, et al., 2019), and protein sequences from Uniprot (UniProt, 2021). Protein structures are from AlphaFold2, with ESMFold (Hie, et al., 2022) used for missing predictions. GO annotations were downloaded from the GOA database (Huntley, et al., 2015), filtered by species and evidence codes, with annotation details recorded. To ensure the reliability of the function labels, we only retained protein annotations that were manually curated by experts and experimentally validated. Specific evidence codes and related experiments are shown in Supplementary Data. All GO annotations follow the Gene Ontology true path rule. We retrieved the go-basic version of the GO label set and, using "is a" and "part of" relationships, traced upwards from existing protein labels to complete each protein's GO label set. The number of labels in each branch is shown in the Table 1. Assuming the number of labels is c and the number of protein samples is N, we represent the protein label vectors in the form of a binary matrix $Y \in (0,1)^{N \times c}$, where 1 indicates the protein possesses the label and 0 indicates it does not. The Gene Ontology can be viewed as comprising three independent branches. Therefore, we classified the proteins according to the branch of the GO label they belong to.

**Table 1** Statistics of the number of labels in each Gene Ontology branch

| Ontology | Label quantity |
|---|---|
| BPO | 19323 |
| MFO | 6445 |
| CCO | 2645 |

Dividing the dataset temporally helps better assess the model's ability to predict protein functions. The training set includes proteins annotated before January 2021, the validation set covers those first annotated between January 2021 and July 2022, and the test set consists of proteins first annotated between August 2022 and August 2023.

### 2.3 Feature extraction stage

#### 2.3.1 Protein sequence representation

We use the ESM-2-650M model to extract features from protein sequences. We input the protein sequences into the ESM-2 model to obtain the amino acid representations from the final layer. By simply averaging these amino acid representations, we derive the protein-level vector representation. The final sequence feature is represented as the vector $H_{se} \in R^{N \times d_1}$, where $d_1$ is the feature dimension of the protein sequence.

#### 2.3.2 Amino acid representation in protein structure

The raw data for protein structure consists of the three-dimensional atomic coordinates of amino acid sequences. We construct the protein structure contact map by connecting amino acids with edges if their Cα atoms are within 10 Å. To derive initial amino acid node embeddings, we employ the node2vec (Grover and Leskovec, 2016) algorithm, which enhances DeepWalk (Perozzi, et al., 2014) with biased random walks. In practice, it uses two hyperparameters $p$ and $q$ to influence the probability of selecting the next hop node. It calculates the selection probabilities for all neighboring nodes of the current node and chooses the node with the highest probability as the next node. Assuming the current node is $v$, the probability of selecting $x$ as the next node is:

$$P(n_i = x | n_{i-1} = v) = \begin{cases} \frac{\pi_{vx}}{Z}, & if (v,x) \in E \\ 0, & otherwise \end{cases} \quad (1)$$

where $\pi_{vx}$ is the transition probability from vertex $v$ to vertex $x$, and Z is a normalization constant. The value of $\pi_{vx}$ is influenced by weights and

hyperparameters. Assuming the walk reaches node $v$ via node $t$, $\pi_{vx}$ is calculated by:

$$\pi_{vx} = \alpha_{pq}(t,x) \cdot w_{vx} \quad (2)$$

$$\alpha_{pq}(t,x) = \begin{cases} \frac{1}{p} & if\ d_{tx} = 0 \\ 1 & if\ d_{tx} = 1 \\ \frac{1}{q} & if\ d_{tx} = 2 \end{cases} \quad (3)$$

where $w_{vx}$ is the weight of the edge between $v$ and $x$, $p$ and $q$ are the hyperparameters introduced by the node2vec algorithm, and $d_{tx}$ represents the distance between node $x$ and node $t$.

Node2vec performs random walks on all nodes in the contact map to generate initial embeddings for each amino acid node. Finally, these embeddings are concatenated with the one-hot encoded vectors of the amino acids.

**2.3.3 Protein structural representation**

Since the heterogeneous networks under study have high degrees, aggregating features from thousands of proteins in a single neighborhood during downstream network propagation tasks imposes significant time and resource constraints. Therefore, we cannot directly input the contact map into the network propagation layer. Inspired by Struct2GO, we employ a structural model train the contact map and extract hidden vectors as low-dimensional structural features. The model performs pooling operations on subgraphs composed of nodes that have the most significant impact on structural features, retaining the most prominent structural characteristics and filtering out noise.

The structural model comprises multiple graph convolution pooling modules. These modules first perform graph convolution on the contact map and amino acid features to aggregate neighborhood information:

$$H^{(l+1)} = \sigma(\widehat{D}^{-\frac{1}{2}}\hat{A}\widehat{D}^{-\frac{1}{2}}H^{(l)}\Theta) \quad (4)$$

where $H^{(l)} \in R^{n \times d_2}$ represents the feature vectors of amino acids at the $l$-th layer, $n$ denotes the number of amino acids in the contact map, and $d_2$ is the dimension of an amino acid vector. $\hat{A}$ is the adjacency matrix of the contact map with self-loops, $\widehat{D}$ is the degree matrix of $\hat{A}$, $\Theta$ represents the learnable parameters of the convolution layer, and $\sigma$ is the activation function. Assuming there are $L$ graph convolution layers, the output of the final graph convolution layer is $H^{(L+1)}$.

Additionally, a single graph convolution layer maps the node features to a one-dimensional space, obtaining the attention scores for each node. Nodes with higher attention scores are retained, with the number of retained nodes determined by the graph pooling rate $k$. Based on the attention scores, the features from the graph convolution are aggregated again to form a pooled subgraph:

$$S = \sigma(\widehat{D}^{-\frac{1}{2}}\hat{A}\widehat{D}^{-\frac{1}{2}}H^{(L+1)}\Theta_s) \quad (5)$$

$$idx = TopSelect(S, k) \quad (6)$$

$$H_{sub} = H_{idx}^{(L+1)} \odot S_{idx} \quad A_{sub} = A_{idx,idx} \quad (7)$$

where $S \in R^n$ represents the attention scores of the amino acids, $idx$ denotes the retained nodes, $H_{sub}$ is the amino acid vector of the pooled subgraph, $A_{sub}$ is the adjacency matrix of the pooled subgraph.

Finally, the module performs graph pooling on the pooled subgraph. We choose to concatenate the results of max-pooling and sum-pooling:

$$H_{out} = \frac{1}{n'}\sum_{i=1}^{n'} X_i \ \| \ max\ X_i \quad (8)$$

where $X = H_{sub}$, $H_{out} \in R^{2d_2}$ is the output of the graph convolution pooling module, $\|$ denotes concatenation, and $n'$ is the number of nodes in the pooled subgraph.

We stack multiple graph convolution pooling modules, continuously condensing the pooled subgraphs to focus the attention on critical amino acid features. Through this method, we learn effective structural features, improving the quality of structural features in downstream tasks. The outputs of multiple layers are summed and input into an MLP multi-label classifier for graph classification, with the goal of predicting GO labels. Binary cross-entropy loss is used to guide backpropagation. Extracting and concatenating the hidden vectors from each structural contact map before the MLP yields the structural features for downstream tasks, $H_{st} \in R^{N \times 2d_2}$.

**2.4 Training stage**

Inspired by PSPGO, our multi-species protein prediction model (hereinafter referred to as the prediction model) employs the same type of heterogeneous network: one part based on functional expression PPI networks and the other based on sequence homology similarity networks. Multi-species PPI networks can be directly downloaded from the STRING database, while homology similarity networks are constructed using the Diamond tool (Buchfink, et al., 2021) to search for similarities between protein sequences. Consequently, we obtain a cross-species heterogeneous network $G = (V, E_1, E_2)$. The advantage of this heterogeneous network is that the underlying homology similarity network provides essential attributes for the proteins, while the surface PPI network supplements functional expression information for the proteins.

**2.4.1 Input layer**

The input to the prediction model is the concatenation of the protein sequence features $H_{se}$ and the protein structural features $H_{st}$:

$$H = H_{se} \ \| \ H_{st} \quad (9)$$

where $H \in R^{N \times (d_1 + 2d_2)}$ represents the concatenated node features. To save time and memory, $H$ needs to be mapped to a lower-dimensional vector space, followed by further mapping using an MLP to avoid issues such as over-smoothing during subsequent convolution processes:

$$H_{mlp}^{(1)} = \sigma(HW_e + b_e) \quad (10)$$

$$H_{mlp}^{(l+1)} = \sigma(LN(H_{mlp}^{(l)}W_{mlp}^{(l)} + b_{mlp}^{(l)})) \quad (11)$$

where $\sigma$ represents the non-linear activation function, $W_e$ and $b_e$ are the learnable parameters of the linear layer, $LN(\cdot)$ is the layer normalization function, $W_{mlp}^{(l)}$ and $b_{mlp}^{(l)}$ are the learnable parameters of the $l$-th layer in the MLP, $H_{mlp}^{(l+1)} \in R^{d_3}$ is the output of the $l$-th layer of the MLP.

**2.4.2 Network propagation layer**

The PPI network is represented as $G_p = (V, E_1)$, and the homology similarity network is represented as $G_s = (V, E_2)$. This study uses a graph attention mechanism to adjust the propagation strength between network nodes, learning an attention score for each edge of the nodes:

$$\alpha_{uv} = \frac{exp(a^T\sigma([W_th_u \| W_th_v]))}{\sum_{i \in N_u} exp(a^T\sigma([W_th_u \| W_th_i]))} \quad (12)$$

where $\alpha_{uv}$ represents the attention score of protein node $u$ to node $v$, $N_u$ denotes the set of neighbors of $u$, $a^T$ and $W_t$ are the learnable weights, $\cdot^T$ denotes the transpose operation, $h_u$ is the node feature of protein $u$. In our experiments, the activation function $\sigma$ is chosen to be the LeakyReLU function.

After updating the weights between nodes in $G_p$ and $G_s$ using the same graph attention parameters, the weights are represented as $A_p$ and $A_s \in R^{n \times n}$, respectively. To propagate the protein node features through the heterogeneous network and avoid overfitting, we use the same network propagation parameters to update the nodes in both networks separately:

$$H_p^{(l)} = \sigma(A_p H^{(l)} W^{(l)} + H^{(l)}) \quad (13)$$

$$H_s^{(l)} = \sigma(A_s H^{(l)} W^{(l)} + H^{(l)}) \quad (14)$$

$$H^{(l+1)} = \sigma(H_p^{(l)} + H_s^{(l)}) \quad (15)$$

where $H^{(l)}$ is the input to the $l$-th layer of network propagation, $W^{(l)}$ is the weight parameter for the $l$-th layer of network propagation, and $H_p^{(l)}$ and $H_s^{(l)}$ are the results of the $l$-th layer propagation in the PPI network and homology similarity network, respectively. The output of each layer of network propagation combines the propagation results of both networks.

### 2.4.3 Output layer and loss function

After sufficient network propagation, the nodes have learned network information from different expression layers. Next, a linear layer and a sigmoid function are used for mapping the node features to the output vector space. Finally, the binary cross-entropy loss between the output probabilities and the ground truth is calculated, and this loss is used to update the parameters through backpropagation:

$$\hat{Y}_{out} = sigmoid(H_{out} W_{out} + b_{out}) \quad (16)$$

$$Loss = -\sum (Y \odot log(\hat{Y}_{out}) + (1-Y) \odot log(1-\hat{Y}_{out})) \quad (17)$$

where $H_{out}$ is the output of the final layer of network propagation, $\hat{Y}_{out} \in R^{n \times c}$ represents the node features mapped to the label vector space, $W_{out}$ and $b_{out}$ are the parameters to be learned for this linear mapping layer, $Y$ represents the ground truth label matrix of the proteins, and $\odot$ denotes element-wise multiplication.

## 2.5 Prediction stage

The network propagation differs between the prediction and training phases. To save resources, only protein feature representations are propagated during training. However, rare GO labels may be overshadowed by dominant features, making feature representation insufficient. In the prediction phase, GO label propagation is added, enabling unlabeled nodes to acquire labels from neighboring nodes. The formula for the label propagation process is as follows:

$$\hat{Y}_p^{(l)} = A_p \hat{Y}^{(l)} \qquad \hat{Y}_s^{(l)} = A_s \hat{Y}^{(l)} \quad (18)$$

$$\hat{Y}^{(l+1)} = Norm(\hat{Y}_p^{(l)} + \hat{Y}_s^{(l)}) \quad (19)$$

where $\hat{Y}^{(l)}$ is the input to the $l$-th layer of network propagation. Specifically, the label input for the first layer of network propagation is $\hat{Y}^{(1)} = Y$, $\hat{Y}_p^{(l)}$ and $\hat{Y}_s^{(l)}$ are the label propagation results on the $l$-th layer of the PPI network and the homology similarity network, respectively. $Norm(\cdot)$ denotes L2 regularization.

The label propagation process is similar to the protein feature propagation process, but it only updates the vectors for proteins without GO labels, i.e., the proteins in the test set. The label vectors updated through network propagation are denoted as $\hat{Y}_{label}$.

The output in the prediction phase is a linear combination of $\hat{Y}_{label}$ and $\hat{Y}_{out}$:

$$\hat{Y} = \varphi \cdot \hat{Y}_{out} + (1-\varphi) \cdot \hat{Y}_{label} \quad (20)$$

where $\varphi \in (0,1)$ is a parameter used to measure the importance of the label vector and protein features, with a step size of 0.1. Based on experimental results, different Gene Ontology branches may have different $\varphi$ value. $\hat{Y} \in R^{N \times c}$ represents the final GO prediction probabilities, with $c$ being the number of labels in the Gene Ontology. $\hat{Y}_{ij} \in [0,1]$ represents the probability that protein $i$ has label $j$.

## 3 Experiments and results

### 3.1 Performance evaluation metrics

Our study focuses on the protein multi-label classification task. Considering the label imbalance problem, we primarily use two evaluation metrics: $F_{max}$ and $S_{min}$.

$F_{max}$ is the highest F1−score among all thresholds $\tau$, which takes into account both precision and recall. $S_{min}$ is a metric specified in CAFA to represent the semantic distance between true functional labels and predicted functional labels. Generally, a larger $F_{max}$ value and a smaller $S_{min}$ value indicate better prediction performance. Additionally, we calculated the Area Under the Precision-Recall Curve (AUPR), which is typically used for datasets with imbalanced positive and negative samples. We provide the calculation formulas for these metrics in Supplementary Data.

In addition, we noticed that weighted $F_{max}$ and weighted AUPR can be used to verify the model's predictive ability for certain important labels. Just like InterLabelGO+ (Liu, et al., 2024) uses Information Context (IC) as a weight, we also use the IC value as the weight of each label to calculate the weighted precision and recall, and then calculate the weighted Fmax and weighted AUPR. The calculation formula of the weighted metrics is also given in the supplementary data.

### 3.2 Experimental settings

The labels of different Gene Ontology branches are independent. To avoid noise, we trained separate models for each of the three Gene Ontology branches to predict their respective labels. The following are the parameter settings used in the experiments.

The dimension of the sequence features $d_1$ is 1280, and the dimension of the structural features $d_2$ is 512. In the structural model, the graph convolution pooling module is set to two layers, each with three graph convolution layers ($L=3$), and the graph pooling rate $k$ is 0.75. The Adam optimizer is used to optimize the parameters, with a learning rate of 5e-4. A three-layer MLP is used for graph classification, and to avoid overfitting, each MLP layer is followed by a dropout layer with a rate of 0.5. The hidden vector dimension $d_3$ in the prediction model is 512, and a single-layer MLP is used for further mapping of the protein representations. The network propagation layer is set to two layers. The prediction model also uses the Adam optimizer to optimize the parameters, with a learning rate of 0.001. In the experiments, the structural model was trained for 20 epochs, and the prediction model was trained for 10 epochs. The linear combination parameter $\varphi$ in the prediction phase takes different values for the three branches: 0.4 in MFO, 0.2 in BPO, and 0.5 in CCO.

**Table 2** Results of comparative experiments.

| Model | BPO | | | MFO | | | CCO | | |
|---|---|---|---|---|---|---|---|---|---|
| | Fmax | Smin | AUPR | Fmax | Smin | AUPR | Fmax | Smin | AUPR |
| SPROF-GO | 0.5054 | 10.0188 | 0.4718 | <u>0.7598</u> | <u>2.7121</u> | <u>0.7935</u> | <u>0.7806</u> | <u>3.3077</u> | <u>0.8162</u> |
| DeepGraphGO+ESM-2 | 0.3078 | 11.9590 | 0.2470 | 0.2615 | 8.5488 | 0.1581 | 0.5640 | 4.9785 | 0.6262 |
| PSPGO+ESM-2 | <u>0.6289</u> | <u>7.5075</u> | <u>0.6582</u> | 0.6395 | 5.2054 | 0.6158 | 0.6758 | 4.0283 | 0.7490 |
| MSNGO+ESM-2 | **0.7332** | **5.9886** | **0.7485** | **0.8102** | **2.4173** | **0.8100** | **0.7920** | **2.8478** | **0.8640** |
| DeepGraphGO+Interproscan | 0.8029 | **4.1623** | 0.8020 | **0.8992** | **1.3749** | **0.9540** | **0.8621** | <u>2.0036</u> | <u>0.9169</u> |
| PSPGO+Interproscan | 0.7949 | 4.3989 | 0.8266 | 0.8750 | <u>1.4203</u> | <u>0.8418</u> | 0.6758 | 4.0283 | 0.7490 |
| MSNGO+Interproscan | **0.8041** | <u>4.3052</u> | **0.8425** | <u>0.8821</u> | 1.4438 | 0.8367 | <u>0.8590</u> | **1.9969** | **0.9259** |

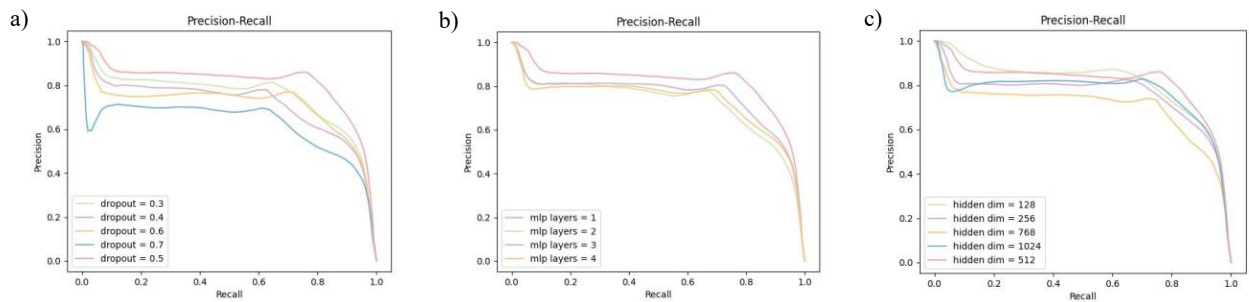

**Fig. 2. PR curves for different hyperparameter values of the prediction model.** a) shows the PR curves for different MLP dropout rates. b) compares PR curves for different numbers of MLP layers. c) compares PR curves for different dimensions of the hidden vectors. In all three figures, the red curve represents the parameter value with the maximum AUPR.

### 3.3 Comparative study

To assess the effectiveness of our proposed method, we conducted comparative experiments against three other multi-species approaches: DeepGraphGO, SPROF-GO, and PSPGO, each of which has its own strengths and weaknesses. DeepGraphGO achieved good results using a simple network propagation approach, with the greatest contribution coming from the protein sequence features extracted by Interproscan (Jones, et al., 2014), which include information such as family domains and functional sites. The quality of the sequence features has a significant impact on the model's performance. In our experiments, we used an alternative method to obtain sequence features to verify this point. SPROF-GO provides pre-trained models for fast GO annotation, offering versatility but being limited by noise in the homology network. PSPGO reduces this noise using a heterogeneous network, yielding good classification results. But it also relies on sequence features from Interproscan, which can limit performance based on extraction methods and quality.

Table 2 displays the results of the comparative experiments, highlighting each method's performance in terms of $F_{max}$, $S_{min}$, and AUPR. The best scores are highlighted in bold, while the second-best scores are underlined. We conducted experiments using two different sequence extraction tools, ESM-2 and Interproscan, to compare the methods. Since the SPROF-GO method includes processing for sequence features, we did not alter its feature input. In the comparative experiments based on ESM-2 sequence features, it is clear that DeepGraphGO performed poorly due to overly simplistic network propagation, which only allows features to propagate within a single PPI network. Although SPROF-GO and PSPGO performed well, they only considered the propagation of sequence features, which introduces certain limitations. The model proposed in this paper outperforms the other three sequence-based methods across all three branches. This superiority is due to our approach, which not only considers sequence features but also introduces structural features into multi-species network propagation for the first time. We designed an effective network propagation algorithm, achieving the best results. Specifically, our model outperforms the second-best method by 10.43%, 5.04%, and 1.14% in the BPO, MFO, and CCO branches, respectively. The outstanding performance in the BPO branch demonstrates our model's stronger classification ability in multi-label classification tasks with many categories. The comparison results using weighted Fmax and weighted AUPR are shown in Table S4 in the Supplementary Data. Compared with Table 2, we found that although the evaluation results of all methods have slightly decreased, the effect of MSNGO is still the best among all methods, which proves that MSNGO also has a good prediction effect on some key proteins.

In the experiments based on Interproscan sequence features, the scores of all methods improved significantly. The most notable improvement was seen in the DeepGraphGO model, which achieved the best results on five metrics. This verifies that DeepGraphGO's performance is highly dependent on the quality of sequence features, making it the most affected by changes in sequence features among the three methods. It is not suitable for the propagation and prediction of simple features, demonstrating poor robustness. Our proposed model still achieved the best results on two metrics in the BPO and CCO branches and ranked first or second on seven metrics. Compared to the results of the experiments using ESM-2, the change in our model's metrics was the smallest, indicating better adaptability to different sequence features. This shows that our model performs well even when using simple, rapidly extracted features. Although the Interproscan tool can extract high-quality sequence features, it is too slow, making it inefficient for large-scale multi-species tasks. Thus, our model clearly exhibits better generalizability.

In order to verify that MSNGO based on multi-species data can enhance the prediction ability on a single-species dataset by aggregating the functional information of similar species, we used MSNGO and Struct2GO to make predictions on Homo sapiens and Saccharomyces cerevisiae, respectively. The specific experimental results are shown in Table S5 in the Supplementary Data. According to the experimental results, the performance of MSNGO on the datasets of two commonly used species is better than that of the single-species Struct2GO model, indicating that the good prediction ability of MSNGO on multiple species can be generalized to a single species. We analyzed that the reason for this advantage may be that MSNGO can aggregate the functional information of similar proteins in other species to improve the prediction probability of potential functions of the current protein.

Overall, our model achieves excellent results in terms of classification performance, robustness, and generalizability.

### 3.4 Ablation study

To validate the effectiveness and importance of each component in our model, we designed four ablation experiments as follows:

(1) **Without Structural Features**: The input to the prediction model retains only the sequence features.

(2) **Without the Structural Model**: The protein contact map's node features obtained from node2vec are aggregated using a mean operation, using the mean as the protein's structural feature without further embedding through the structural model.

(3) **Without the Network Propagation Layer**: The extracted structural and sequence features are directly input into the MLP classifier in the prediction model to obtain the final probabilities.

(4) **Without Label Propagation**: During the prediction phase, only the outputs of the protein features are used as the final probabilities.

The results of the ablation experiments are shown in the Table 3. It can be observed that the absence of the network propagation layer leads to the most significant drop in metrics, demonstrating the effectiveness of the network propagation algorithm in the prediction model.

**Table 3.** Ablation experiment results on MFO

| Methods | Fmax | Smin | AUPR |
| --- | --- | --- | --- |
| without structural feature | 0.6395 | 5.2054 | 0.6158 |
| without the structural model | 0.6609 | 4.8855 | 0.6299 |
| without the network propagation layer | 0.5449 | 5.2244 | 0.5255 |
| without label propagation | 0.7338 | 3.3048 | 0.7910 |
| MSNGO | **0.8102** | **2.4173** | **0.8100** |

In the ablation experiment, structural features are the second most important contributor to the improvement of model performance. When only sequence features are used without structural features, the model score $F_{max}$ drops significantly by 0.17, indicating that the introduction of structural features effectively improves classification performance. This result is particularly important because similar sequences may form different structures, which may lead to significant changes in function. Therefore, it is difficult to fully analyze the complexity of protein function by relying solely on sequence information, while structural features can capture three-dimensional conformations, functional domain arrangements, and potential functional sites that are difficult to reveal with sequence information. By integrating structural features, the model can better distinguish these complex situations, thereby performing more accurately in function prediction.

Additionally, the ablation experiments for the structural model and label propagation further confirmed the importance of the structural model in extracting structural features and the supplementary role of label propagation in enhancing the prediction results.

### 3.5 Parameter sensitivity analysis

We observed the impact of hyperparameters such as dropout, the number of MLP layers, and the hidden vector dimension on the performance of the prediction model. Using the controlled variable method, we varied these parameters one at a time and obtained the PR curves for different values. We evaluated the impact of each parameter on model performance by comparing the area under the PR curve. The optimal parameter values were then selected for practical application. The range of these parameter values is shown in Table 4. The PR curves, shown in Figure 2, illustrate the changes in PR for different values of dropout, the number of MLP layers, and the hidden vector dimension.

**Table 4.** Range of values for hyperparameter analysis

| Hyperparameter | Range |
| --- | --- |
| MLP dropout | 0.3、0.4、0.5、0.6、0.7 |
| number of MLP layers | 1、2、3、4 |
| hidden dimension | 128、256、512、768、1024 |

The dropout in the MLP is applied after each layer of the neural network, which helps to prevent overfitting by deactivating a portion of the neurons. Experiments show that a dropout rate of 0.5 yields the largest PR curve area, while 0.7 results in the smallest, suggesting that higher dropout rates cause information loss and reduce performance. Therefore, a 0.5 dropout rate is most effective.

We aim to use an MLP in the prediction model to further map the protein representations, thereby avoiding issues such as over-smoothing that can arise during network propagation. According to experimental results, the number of MLP layers should be kept small. When the number of MLP layers increases, the neural network becomes deeper, which can lead to overfitting and reduce the model's performance. Therefore, to address the overfitting issues associated with deep MLP networks, we set the number of MLP layers to just 1.

Setting the hidden vector dimension to 512 in the model is optimal. This is because larger dimensions would lead to wasted memory space with minimal gain, while smaller dimensions would fail to fully represent the protein features, thereby affecting the model's classification capability.

In addition, we show the discussion of different values of the graph pooling rate k in Table S6 in the Supplementary Data. We set k to 0.25, 0.5, 0.75, and 0.8 respectively. The experimental results show that when k is 0.75, both weighted Fmax and weighted AUPR can achieve the best. In the process of increasing from 0.25 to 0.75, the Fmax value rises slowly.

From 0.75 to 0.8, the model has a significant decrease in all three indicators. Therefore, we selected 0.75 as the fixed value of k in the experiment.

## 4  Conclusion

In this paper, we propose a multi-species protein function prediction model called MSNGO, based on structural features and heterogeneous network propagation. This model combines protein sequence and structural features, integrating the advantages of PPI networks and homology similarity networks, effectively predicting protein functions. Specifically, we use the protein structures predicted by AlphaFold2 and train a structural model to extract protein-level representations. After combining sequence features from ESM-2, protein representations are propagated and updated in heterogeneous neural networks as node vectors. The method proposed in this paper introduces protein structural features into the multi-species domain for the first time. The high-precision structural predictions of AlphaFold2 facilitate the acquisition of structural features, making the protein representations more comprehensive by including both structural and sequence features. This approach addresses the limitations of relying solely on sequence features and effectively improves the accuracy of multi-species protein function prediction. Comparative experimental results indicate that MSNGO achieves state-of-the-art performance, demonstrating excellent robustness and generalizability in the field of multi-species protein function prediction.

In our future work, we will continue to explore new methods for multi-species protein function prediction. We have observed that existing protein function prediction methods adopting multimodal approaches have achieved excellent results. Therefore, in future research, we plan to improve data integration methods and further enhance prediction performance by leveraging multimodal fusion strategies.

Additionally, we can make improvements to the PPI network. The MSNGO method primarily relies on the homology similarity network within the heterogeneous network for cross-species label propagation. However, the PPI network can also be aligned to identify similar network structures, enabling cross-species label propagation directly within the PPI network. Effectively utilizing multi-species PPI network information can further enhance prediction performance.

## Fundings

This work was supported by the grants from the National Key R&D Program of China (2021YFA0910700), National Natural Science Foundation of China (32470704), Guangdong Provincial Key Laboratory of Novel Security Intelligence Technologies (2022B1212010005).

## References

Ashburner, M., *et al.* (2000) Gene ontology: tool for the unification of biology, *Nature genetics*, **25**, 25-29.

Barot, M., *et al.* (2021) NetQuilt: deep multispecies network-based protein function prediction using homology-informed network similarity, *Bioinformatics*, **37**, 2414-2422.

Buchfink, B., Reuter, K. and Drost, H.-G. (2021) Sensitive protein alignments at tree-of-life scale using DIAMOND, *Nature methods*, **18**, 366-368.

Dessimoz, C., Škunca, N. and Thomas, P.D. (2013) CAFA and the open world of protein function predictions, *Trends in Genetics*, **29**, 609-610.

Gligorijevic, V., *et al.* (2021) Structure-based protein function prediction using graph convolutional networks, *Nat Commun*, **12**, 3168.

Grover, A. and Leskovec, J. (2016) node2vec: Scalable Feature Learning for Networks, *KDD*, **2016**, 855-864.

Hie, B., *et al.* (2022) A high-level programming language for generative protein design, *bioRxiv*, 2022.2012.2021.521526.

Huntley, R.P., *et al.* (2015) The GOA database: Gene Ontology annotation updates for 2015, *Nucleic Acids Research*, **43**, D1057-D1063.

Jiao, P., *et al.* (2023) Struct2GO: protein function prediction based on graph pooling algorithm and AlphaFold2 structure information, *Bioinformatics*, **39**, btad637.

Jing, B., *et al.* (2020) Learning from protein structure with geometric vector perceptrons. *International Conference on Learning Representations*.

Jones, P., *et al.* (2014) InterProScan 5: genome-scale protein function classification, *Bioinformatics*, **30**, 1236-1240.

Jumper, J., *et al.* (2021) Highly accurate protein structure prediction with AlphaFold, *Nature*, **596**, 583-589.

Kipf, T.N. and Welling, M. (2016) Semi-supervised classification with graph convolutional networks, *arXiv preprint arXiv:1609.02907*.

Liu, Q., Zhang, C. and Freddolino, L. (2024) InterLabelGO+: unraveling label correlations in protein function prediction, *Bioinformatics*, **40**.

Ma, W., *et al.* (2024) Annotating protein functions via fusing multiple biological modalities, *Communications Biology*, **7**, 1705.

Perozzi, B., Al-Rfou, R. and Skiena, S. (2014) Deepwalk: Online learning of social representations. *Proceedings of the 20th ACM SIGKDD international conference on Knowledge discovery and data mining.* pp. 701-710.

Rives, A., *et al.* (2021) Biological structure and function emerge from scaling unsupervised learning to 250 million protein sequences, *Proceedings of the National Academy of Sciences*, **118**, e2016239118.

Singh, R., Xu, J. and Berger, B. (2008) Global alignment of multiple protein interaction networks with application to functional orthology detection, *Proc Natl Acad Sci U S A*, **105**, 12763-12768.

Szklarczyk, D., *et al.* (2019) STRING v11: protein-protein association networks with increased coverage, supporting functional discovery in genome-wide experimental datasets, *Nucleic acids research*, **47**, D607-D613.

Torres, M., *et al.* (2021) Protein function prediction for newly sequenced organisms, *Nature Machine Intelligence*, **3**, 1050-1060.

UniProt, C. (2021) UniProt: the universal protein knowledgebase in 2021, *Nucleic Acids Res*, **49**, D480-D489.

Varadi, M., *et al.* (2022) AlphaFold Protein Structure Database: massively expanding the structural coverage of protein-sequence space with high-accuracy models, *Nucleic acids research*, **50**, D439-D444.

Wang, W., *et al.* (2025) DPFunc: accurately predicting protein function via deep learning with domain-guided structure information, *Nat Commun*, **16**, 70.

Wu, K., *et al.* (2022) PSPGO: Cross-species heterogeneous network propagation for protein function prediction, *IEEE/ACM Transactions on Computational Biology and Bioinformatics*.

You, R., *et al.* (2021) DeepGraphGO: graph neural network for large-scale, multispecies protein function prediction, *Bioinformatics*, **37**, i262-i271.

Yuan, Q., *et al.* (2023) Fast and accurate protein function prediction from sequence through pretrained language model and homology-based label diffusion, *Briefings in bioinformatics*, **24**, bbad117.

Zheng, R.T., Huang, Z.J. and Deng, L. (2023) Large-scale predicting protein functions through heterogeneous feature fusion, *Briefings in Bioinformatics*, **24**.